\documentclass[11pt]{article}
\usepackage{acl2020}
\usepackage{times}
\usepackage{url}
\def\aclpaperid{***}
\usepackage{latexsym}
\usepackage{amsmath}
\usepackage{booktabs}
\usepackage{graphicx}
\usepackage{multibib}
\usepackage{subfig}
\usepackage{bbm}
\usepackage{makecell}
\usepackage{tikz}
\usepackage[T1]{fontenc} 
\usepackage{tabularx,pbox}
\usetikzlibrary{fit,positioning}
\usetikzlibrary{calc, automata, chains, arrows.meta}
\usepackage{newfloat}

\DeclareFloatingEnvironment[listname=Box,name=Box]{story}
\expandafter\def\expandafter\UrlBreaks\expandafter{\UrlBreaks
  \do\a\do\b\do\c\do\d\do\e\do\f\do\g\do\h\do\i\do\j%
  \do\k\do\l\do\m\do\n\do\o\do\p\do\q\do\r\do\s\do\t%
  \do\u\do\v\do\w\do\x\do\y\do\z\do\A\do\B\do\C\do\D%
  \do\E\do\F\do\G\do\H\do\I\do\J\do\K\do\L\do\M\do\N%
  \do\O\do\P\do\Q\do\R\do\S\do\T\do\U\do\V\do\W\do\X%
  \do\Y\do\Z}

\aclfinalcopy

\title{``If it Bleeds, it Leads'': A Computational Approach to Covering Crime in Los Angeles}

\author{Divya Choudhary* 
    \qquad Alexander Spangher* \\
  Information Sciences Institue / University of Southern California \thanks{* Authors contributed equally. Author order is based on last-name alphabetical, descending.}\\
  {\tt \{dchoudha, spangher\}@isi.edu} \\
}

\begin{document}
\maketitle
\begin{abstract}
	\textbf{NOTE (2022-06-14): }This work was done for a class project in Fall 2019 and is shown, unless noted, without changes.
	
    Developing and improving computational approaches to covering news can increase journalistic output and improve the way stories are covered. In this work we approach the problem of covering crime stories in Los Angeles. We present a machine-in-the-loop system that covers individual crimes by (1) learning the prototypical coverage archetypes from classical news articles on crime to learn their structure and (2) using output from the Los Angeles Police department to generate ``lede paragraphs'', first structural unit of crime-articles. We introduce a probabilistic graphical model for learning article structure and a rule-based system for generating ledes. We hope our work can lead to systems that use these components together to form the skeletons of news articles covering crime. \footnote{This work was done for a class project in Jonathan May's Advanced Natural Language Processing Course, Fall, 2019.}
\end{abstract}

\section{Introduction}

Newspaper coverage of crime serves several purposes. It memorializes the victims of crimes, it informs the public, and it helps hold officers and city officials accountable for an important dimension of urban health \cite{katz1987makes}. Indeed, crime news is a cornerstone of local and national media coverage: some reports put crime coverage at 50\% of total U.S. news coverage by volume \cite{schildkraut2017crime,crimestat}. 

However, declining newspaper advertisement revenues have had a significant impact on newspapers in the U.S. \cite{pewnewsadvertising}. This has led to severe staff cuts, paper closings and diminishing in journalistic quality, especially at local outlets \cite{freedman2009new}. This disproportionately effects newspapers abilities to cover local crime.

Simultaneously (and orthogonally), much of the criticism of existing crime coverage in news media focuses on the tendency of the press to sensationalize crime \cite{chermak1994body}, to focus on police-officer viewpoints \cite{marsh1991comparative}, and to overreport certain types (and demographics) of crime \cite{antunes1977representation}. Surveys have found that this leads to a misunderstanding of local risks \cite{gilliam1996crime}. Some suggestions for improvement include ``more investigative and analytical coverage'' \cite{cohen1975comparison}.

\begin{figure}
    \centering
    \includegraphics[width=.7\linewidth]{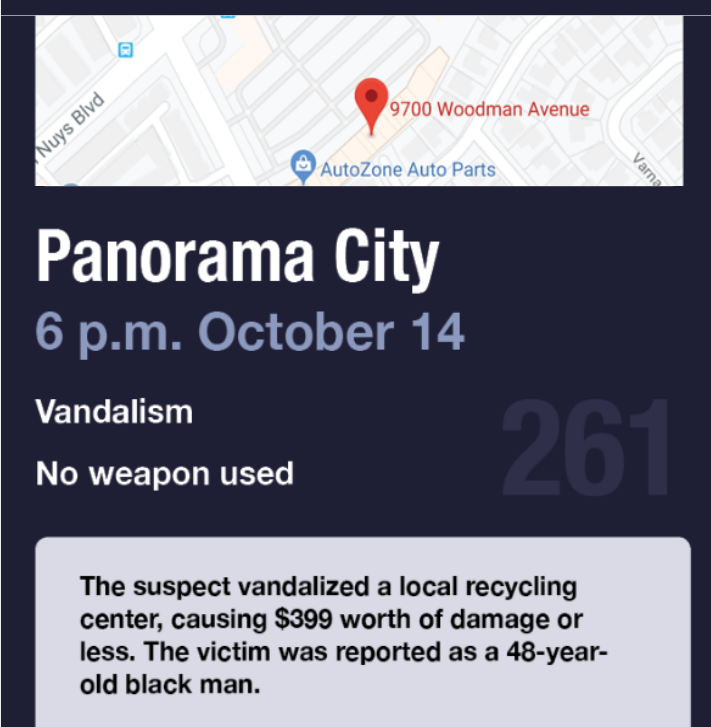}
    \caption{Sample crime template currently generated by \url{xtown.la}.}
    \label{fig:sample-xtown-template}
\end{figure}

Thus, we see a dual purpose in taking computational approaches to covering crime: these approaches can (1)  enhance local journalism's ability to cover more crimes and (2) allow for a more analytical and factual reporting that provides appropriate context.

In this work, we present a machine-in-the-loop system for achieving these dual goals. We start with the archetypical news article as a basis for our endeavors. We propose a two-step approach: first, the initial crime-event is described using a set of templates and data generated by the Los Angeles Police Department (LAPD). Then, appropriate context-paragraphs are recommended by models trained to learn the structure of paragraphs in articles and their flow. We provide the code.\footnote{\url{https://github.com/alex2awesome/lapd-hate-crimes}}\newline

Our contributions are threefold:
\begin{itemize}
    \item We formulate the problem of crime-coverage as well as an approach that puts the human journalist at the center of a machine-in-the-loop system.
    \item We provide a set of templates that translate LAPD crime-codes into lede paragraphs.
    \item We provide a novel graphical model as well as a Gibbs Sampler and code that learns the structure of news articles in terms of paragraph transitions in an unsupervised manner.
\end{itemize}

Taken together, we envision a \textit{computational journalism} system that can bootstrap crime coverage, thus allowing journalists to be more efficient, cover more stories, and do so from a more principled perspective. Our demonstration here shows not only that this is possible, but feasible for one type of crime story. This is an approach that can be adopted and expanded for other types of stories as well. 

\section{Background}

\begin{figure}
    \centering
    \includegraphics[width=.9\linewidth]{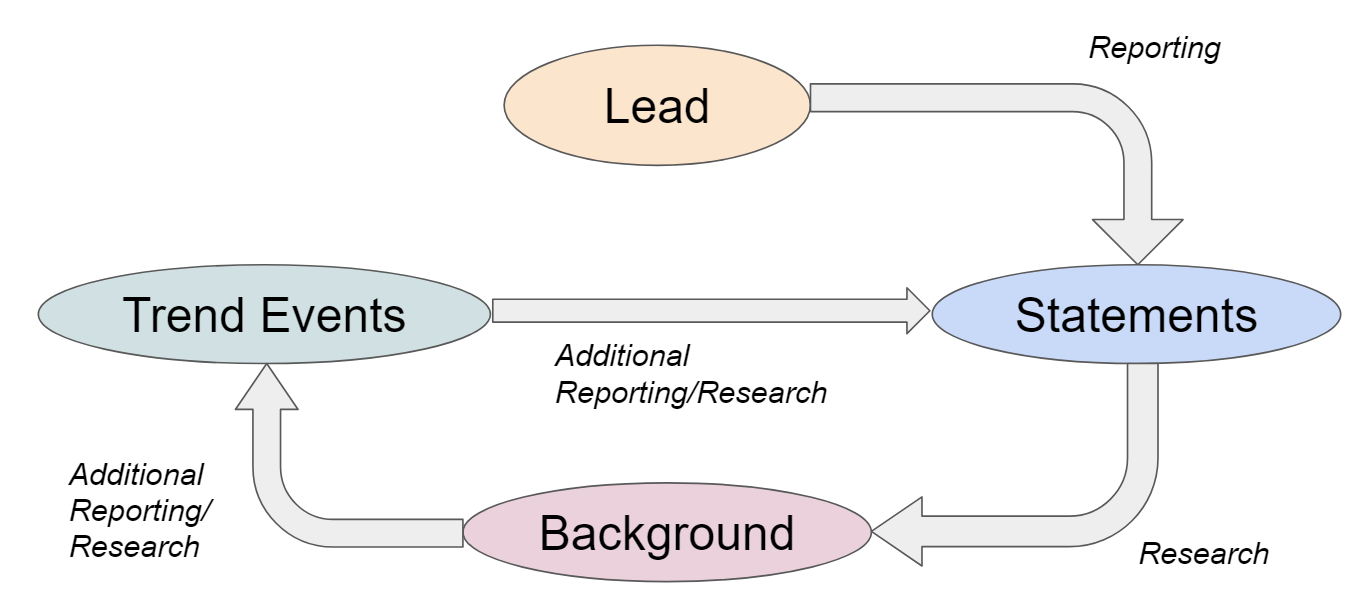}
    \caption{A graphic representation of newsroom}
    \label{fig:newsroomflow}
\end{figure}

The standard flow for writing a crime article is typically simple: (1) a lede is identified, (2) statements are gathered from police, victim (or family), or other associated entities (3) broader context is added via background, which is then (4) connected to a trend. This flow is shown in Figure \ref{fig:newsroomflow}, along with the procedures that are followed by journalists at each step. 

\begin{story}
    \centering
    \begin{tabular}{|p{.7\linewidth}|p{.15\linewidth}|}
    \hline    
      \multicolumn{2}{|l|}{``K-POP STAR IMPRISONED, SEX CRIMES''}\\
      \hline
      \textbf{Paragraph} & \textbf{Type}\\
      \hline
      A South Korean court found two K-pop stars guilty of having sex with a woman without her consent.
      & 1. \underline{\textit{T0:}} \textit{Lede} \\
      
      The court said in a statement that they were convicted of committing ``special quasi-rape''... & 2. \underline{\textit{T2:}} \textit{Statement}\\
      South Korean entertainment industry is hugely popular, but have suffered a series of sexual scandals that has revealed a dark side. & 3. \underline{\textit{T1:}} \textit{Background}\\
      On Sunday, K-pop musician G. Hara was found dead at her home... & 4. \underline{\textit{T:3}} 2ndary event
      \\
    \hline
    \end{tabular}
  \caption{A sample crime news article article hand-labeled to show the transitions between paragraph-types. (Url: \url{https://lat.ms/2PnnNLl})}
  \label{box:sample-article}
\end{story}

We show an example article labeled with this procedural flow in Box \ref{box:sample-article}. As can be seen, the first paragraph describes the main event, the \textit{Lede} (Paragraph Type: 0). Following paragraphs each add different elements to the story: background, statements and secondary events. Together, this conveys to the reader not just the primary event of interest, but where it fits into the broader world of current events and trends.

\section{Problem Statement}

Thus, we wish to break down the process of newswriting by identifying the news article as a set of paragraphs:
\begin{align}
    article = [p_1, p_2, p_3 ...]\\
    type(p_i) = t \in \{1, ..., T\}
\end{align}

As shown (eqn 1), the article is an ordered set of paragraphs, where each paragraph has a type. Our objectives are two-fold:

\begin{enumerate}
    \item We wish to learn the archetypical types of paragraphs in a news article and the transition probabilities between each paragraph.
    \item We wish to take a computational approach to generating different types of paragraphs.
\end{enumerate}

In the current work, we present both our approach to learning these paragraph archetypes and an implementation of generation techniques for the \textit{lede} paragraph for one type of story: crime stories in Los Angeles.

\section{Structure Learning}

\subsection{Generative Model}
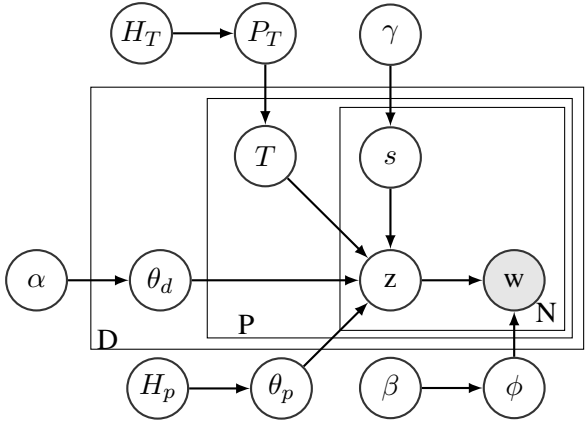
\begin{figure}[t]
\centering
\begin{tikzpicture}
\tikzstyle{main}=[circle, minimum size = 8mm, thick, draw =black!80, node distance = 8mm]
\tikzstyle{connect}=[-latex, thick]
\tikzstyle{box}=[rectangle, draw=black!100]
%
%
\node[main](thetaa) [label=center:$\alpha$] { };
\node[main](thetad) [right=of thetaa, label=center:$\theta_{d}$] { };
\node[main] (z) [right=2.2cm of thetad,label=center:z] {};
\node[main] (T) [above left=1.5 cm of z, label=center:$T$] { };
\node[main] (Pt) [above=of T,label=center:$P_T$] { };
\node[main] (Ht) [left=of Pt,label=center:$H_T$] { };
\node[main] (s) [above=of z,label=center:$s$] { };
\node[main] (gamma) [above=of s,label=center:$\gamma$] { };
\node[main, fill = black!10] (w) [right=of z,label=center:w] { };
\node[main] (Pw) [below=.6cm of w,label=center:$\phi$] { };
\node[main] (Hw) [left=of Pw,label=center:$\beta$] { };
\node[main] (thetap) [below left=1.2cm of z,label=center:$\theta_p$] { };
\node[main] (Hp) [left=of thetap,label=center:$H_p$] { };
%
%
\path
(thetaa) edge [connect] (thetad)
(thetad) edge [connect] (z)
(Pt) edge [connect] (T)
(Ht) edge [connect] (Pt)
(T) edge [connect] (z)
(z) edge [connect] (w)
(Pw) edge [connect] (w)
(Hw) edge [connect] (Pw)
(gamma) edge [connect] (s)
(s) edge [connect] (z)
(Hp) edge [connect] (thetap)
(thetap) edge [connect] (z)
;
%
%
\node[rectangle, inner sep=-2.5mm, fit= (w) (z) (s),label=below right:N, xshift=-.2mm] {};
\node[rectangle, inner sep=2.5mm,draw=black!100, fit=(w) (z) (s)] {};
\node[rectangle, inner sep=-1mm, fit= (T) (w) (s),label=below left:P, xshift=-5mm] {};
\node[rectangle, inner sep=3.5mm,draw=black!100, fit=(T) (w) (s)] {};
\node[rectangle, inner sep=1mm, fit= (thetad) (w) (T),label=below left:D, xshift=-14mm] {};
\node[rectangle, inner sep=5.0mm,draw=black!100, fit=(thetad) (w) (T)] {};
\end{tikzpicture}
\caption{Plate diagram for Source Topic Model}
\label{fig:plateandstick}
\end{figure}

Our generative model for the structure of news articles assumes that each news article is composed of a collection of paragraphs, each of type $T$, which combined with a background document topic serves to generate the words observed in the document. Our goal is to learn the latent variable $T$ associated with each paragraph. Our generative story is as follows:

For paragraph type $p=1,...,P$, sample paragraph-type distribution $\theta_p \sim Dir(H_p)$. 

For document $d=1, ..., D$:
    \begin{enumerate}
    \item Sample background topic distribution $\theta_d \sim Dir(\alpha)$
    \item For paragraph $1,...P_d$:
    \begin{enumerate}
        \item Sample par.-type $T \sim Cat(P_T)$
        \begin{enumerate}
            \item For word $1, ..., N_p$:
            \begin{enumerate}
                \item Sample word-type $s \sim Bin(\gamma)$
                \item if $s=$``paragraph word'', sample word-topic $z \sim Cat(\theta_p^{(T)})$
                \item if $s=$``document word'', sample word-topic $z \sim Cat(\theta_d)$
                \item Sample word $w\sim Cat(\phi^{(z)})$
            \end{enumerate}
        \end{enumerate}
    \end{enumerate}
\end{enumerate}

We show a graphical representation of our model in Figure \ref{fig:plateandstick}. On a high level, we assume each paragraph type shares general linguistic characteristics across documents: for example, a background paragraph might have words like ``percent'' or ``nation''. However, we assume that many words in all paragraphs are document-specific: words like ``gun'' or ``hate'' in an article about hate-crime might occur in all paragraphs.

Our model shares some similarities with \newcite{bamman2013learning}'s Persona Topic Model, which also infers a latent type across documents, a notable departure from the classical Latent-Dirichlet Allocation model \cite{blei2003latent}. \newcite{bamman2013learning}'s work has a similar inference structure, except \newcite{bamman2013learning} observe the switching variable $s$. In our model, we leave this unobserved.

\subsection{Inference}

We construct a joint probability and collapse out the Dirichlet priors, $\theta_d$, $\theta_p$, $P_T$ and $\phi$. We iteratively sample the type of each paragraph $T_p$, and the switching variable and word-topic, $[s_{d, n}, z_{d, n}]$. $s_{d,n}$ and $z_{d,n}$ are sampled as a block. We now describe the inference procedures for each variable.

\noindent\textbf{Paragraph-Type Inference:} First, for each paragraph in each document, we sample a paragraph type $T_p \in \{1, ..., T\}$ according to:
\begin{align*}
    p(T_p &| T_{-p}, z, s, \gamma, H_T, \alpha)\\
    &\propto p(T_p | H_T) \times p(z | s=\text{``par''}, H_p)\\
    &(H_T + c_{T_p, *}^{-p})
    \times 
    \prod_{j \in N_p} \frac{(c_{z_j, T_p, *} + H_p)}{(c_{*, T_p, *} + K H_p)}
\end{align*}

where the first term in the product is the probability associated with each paragraph type: $c_{T_p, *}^{(-p)}$ is the count of all paragraphs of type $T_p$ without considering the current paragraph's assignment. The second term is the probability associated with word topics in the paragraph of type $s=\text{``paragraph''}$. Here, $N_p$ is the set of all word-topics in the current paragraph of this type. $c_{z_j, T_p, *}$ represents the count of all word topics of type $z_j$ associated with each paragraph type $T_p$ and $c_{*, T_p, *}$ represents the count of all word topics associated with paragraph type $T_p$. $K$ is the number of topics.\footnote{We note that this sampling equation is subject to review. We dropped the term $p(s|\gamma)$, as it did not appear to vary in the sampling procedure. However, the equation as it stands appears to violate d-separation, as $s$ is a parent of a child of $T$.}

\noindent\textbf{Switching Variable Inference:} Next, for each word in each paragraph, we sample a switching variable, $s_{d, n}$ which determines if the word contributes to the paragraph-type topic, or the background document topic. We use the following sampling equation:
\begin{align*}
    &p(s_{(d, n)} | s_{-(d, n)}, z, \gamma)\\
    &\propto p(s_{(d, n)} | \gamma) \times p(z_{(d, n)} | \theta_p, \theta_d)\\
    &\propto 
    \begin{cases}
    \gamma \times 
    \frac{(c_{z_{(d, n)}, T_p, *} + H_p)}{(c_{*, T_p, *} + K H_p)}, \text{ if $s_{(d, n)}$ = ``par''}\\
    (1 - \gamma) \times 
    \frac{(c_{z_{(d, n)}, d, *} + H_p)}{(c_{*, d, *} + K H_p)}, \text{ if $s_{(d, n)}$ = ``doc''}
    \end{cases}
\end{align*}

Where the first term in both cases represents the Bernoilli probability term associated with the switching variable. The second term represents the probability that word-topic $k$ is associated with either the current paragraph type or the document. Here, $c_{z_{(d, n)}, T_p, *}$, $c_{*, T_p, *}$ and $K$ are the same as defined in the previous section. $c_{z_{(d, n)}, d, *}$ is the count of all word-topics of type $z_{(d, n)}$ associated with document $d$ and $c_{*, d, *}$ is the count of all word-topics associated with document $d$.

\noindent\textbf{Word-topic Inference:} Finally, for each word-topic in each paragraph, we sample $z_{(d, n)}$ which determines the topic of the word. We use the following sampling equation:
\begin{align*}
    &p(z_{(d, n)} | s_{(d, n)}, \theta_p, \theta_d, \alpha, H_p)\\
    &\propto
    \begin{cases}
    (c_{z_{(d, n)}, T_d, *}^{-(d, n)} + H_p) 
    \times W_{(d,n)}, \text{ if $s_{(d, n)}$ = ``par''}\\
    (c_{z_{(d, n)}, d, *}^{-(d, n)} + \alpha) 
    \times W_{(d,n)}, \text{ if $s_{(d, n)}$ = ``doc''}
    \end{cases}\\
    &\text{where: } W_{(d,n)} = \frac
        {(c_{z_{(d, n)}, *, w_{(d, n)}, *} + \beta)}
        {(c_{z_{(d, n)}, *, *, *} + V \beta)} 
\end{align*}

Where the first term represents the probability associated with each word-topic. In these terms, $c_{z_{(d, n)}, T_d, *}^{-(d, n)}$ and $c_{z_{(d, n)}, d, *}^{-(d, n)}$ are as defined above, except with the current word-topic assignment held out. The second term represents the probability of the word-topic being associated with the word, $w_{(d, n)}$. Here, $c_{z_{(d, n)}, *, w_{(d, n)}, *}$ represents the count of all words $w_{(d, n)}$ associated with topic $z_{(d, n)}$, $c_{z_{(d, n)}, *, *, *}$ represents counts of $z_{(d, n)}$ overall and $V$ represents the size of the vocabulary.

\subsection{Data}

We use the \textit{New York Times} Annotated Corpus\footnote{\url{https://catalog.ldc.upenn.edu/LDC2008T19}} to train our model, which contains 1.8 million articles published during 1987--2007, as well as metadata information for each article, including the date of publication and the page of the newspaper the article was printed on. 

We filter to all articles that have the word ``crime'' in it as a rough heuristic to identify crime articles. Additionally, we also filter to all \textit{New York Times} that were published between Monday-Friday.\footnote{Stories published on weekend days tend to be longer, investigative pieces or analysis pieces that have significantly different structure from typical daily news stories. Thus, to bound our analysis we focus on weekday front-page stories.} This results in approx. $48,000$ articles.\footnote{Because of a technical glitch, we only run our topic model over the first $1,000$ documents.} We remove a set of publication-specific stopwords.

\subsection{Analytical results}

To reiterate the goal of our model, we are interested in modeling the structure of an article by modeling different paragraph types. We hypothesize that the paragraph-type is determined by a set of words common across documents. As such, to explore the performance of our model, we explore (1) the words associated with paragraph-structure (i.e. $w_{d,n} s.t. s_{d,n}=0$) (2) the paragraph types we learn and, (3) transitions between paragraph types and accuracy of tagging. As we lack expert annotation for a thorough evaluation, for (3), we will examine how well our model tags the first paragraph with a consistent type, as this is most likely to be a lede paragraph.

\noindent\textbf{Words associated with Paragraph Structure.} As shown in Table \ref{tbl:top-words-s-1}, the words most associated with the document-specific topic, $s=1$, are topic-specific words, as we'd expect. On the other hand, the words most associated with the paragraph-type topic, $s=0$, appear to be more general descriptive words. 

\begin{figure}
    \centering
    \includegraphics[width=.7\linewidth]{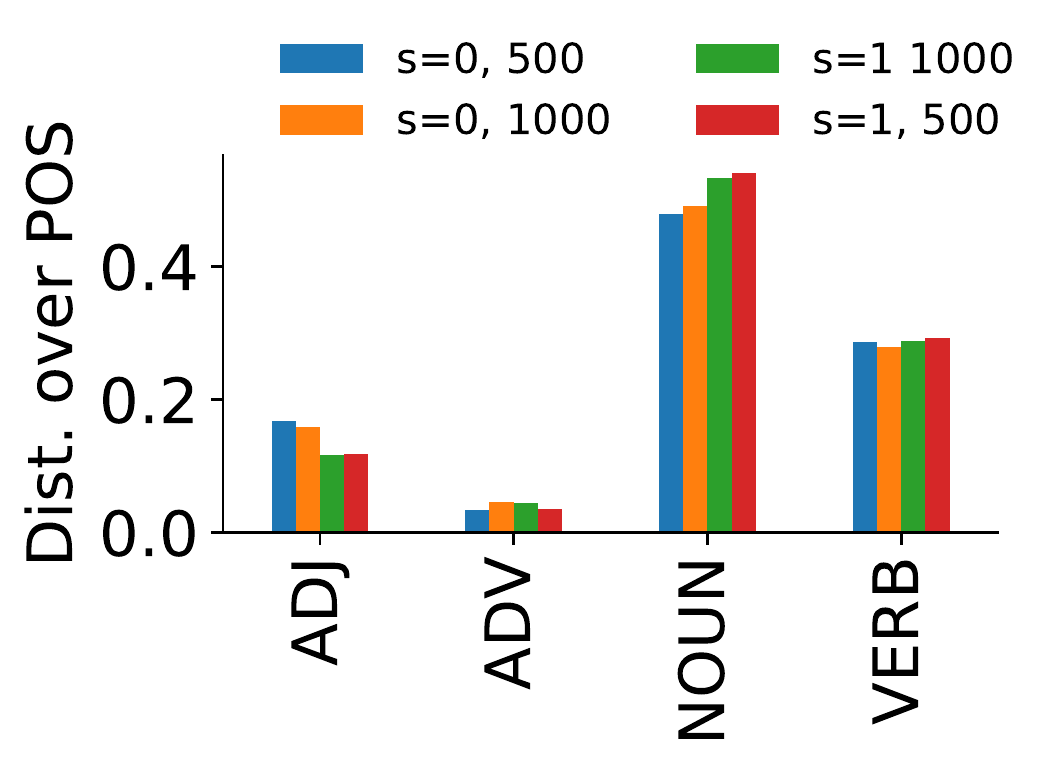}
    \caption{Distribution over POS tags for words most associated with either the paragraph-type ($s=0$) or the background document topic ($s=1$).}
    \label{fig:switchin-var-pos}
\end{figure}

We further examine the nature of these words in Figure \ref{fig:switchin-var-pos}, which shows the distribution over POS tags among the top word for $s=0$ and $s=1$ respectively. As can be seen, the words more associated with $s=0$ are descriptive, ADJ, while the words more associated with $s=1$ are specific, NOUN. The lede paragraph learned by our topic model is most likely to be of type 2, followed by 0 and 4.

\noindent\textbf{Paragraph-Type Words}: We show the top topics associated with each paragraph-type in Table \ref{tbl:partype-topics}. It appears that paragraph-type 6 captures some sense of statement, the words ``charges'' and ``states'' are prominent. On the other hand, paragraph-type 7 seems to be more background: ``percent'' and ``according'' are both prominent, which are typically used to summarize surveys or expert opinions.

\noindent\textbf{Transitions and Lede-Paragraph Identification:} We compare our model with a simple baseline, \textbf{KMeans}: first, we average the glove vectors of all the words in each paragraph, then learn KMeans clusters over each paragraph.

\begin{figure}[t]
    \centering
    \subfloat[Paragraph Topic Model.]{
    \includegraphics[width=.45\linewidth]{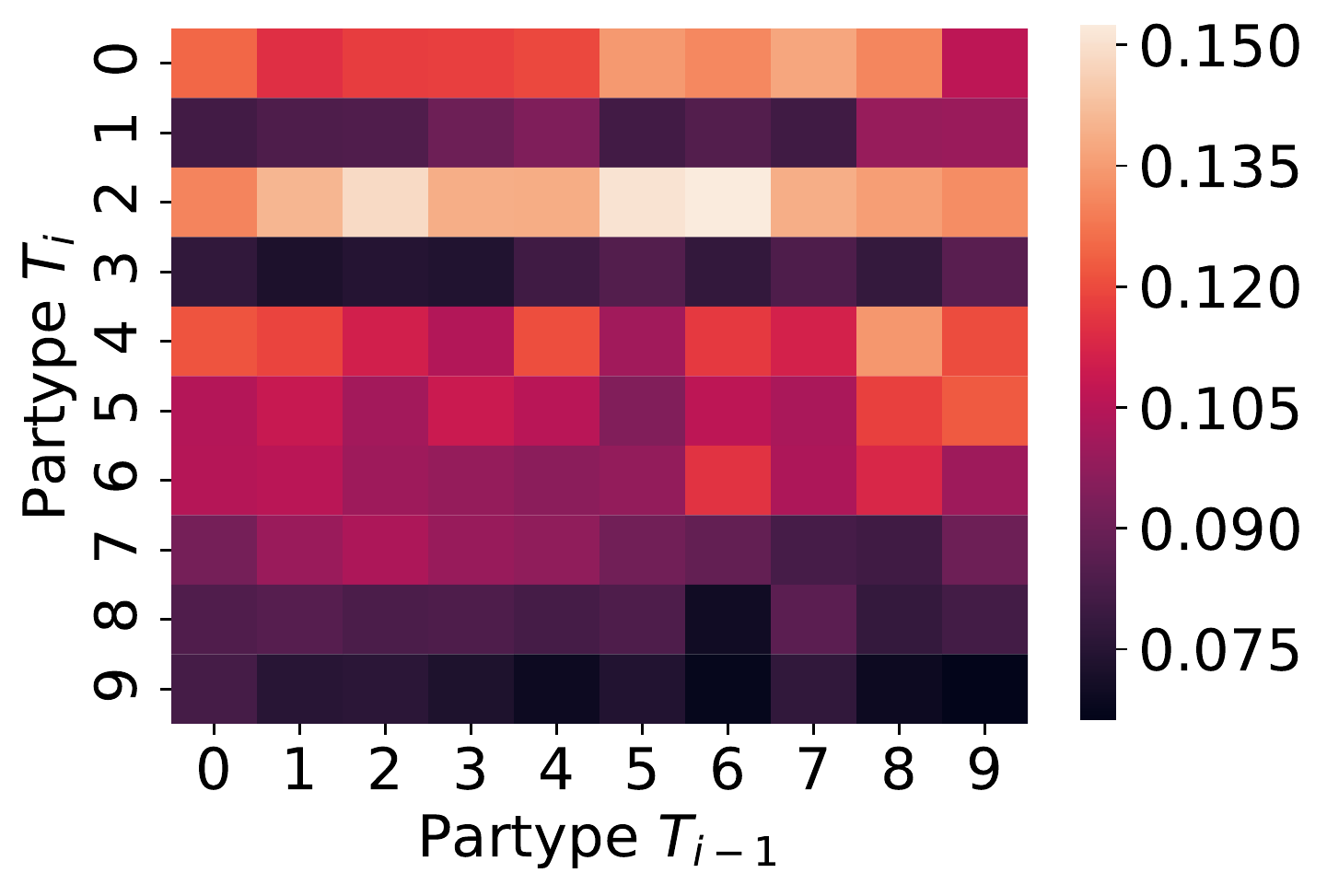}
    }
    \subfloat[KMeans.]{
    \includegraphics[width=.45\linewidth]{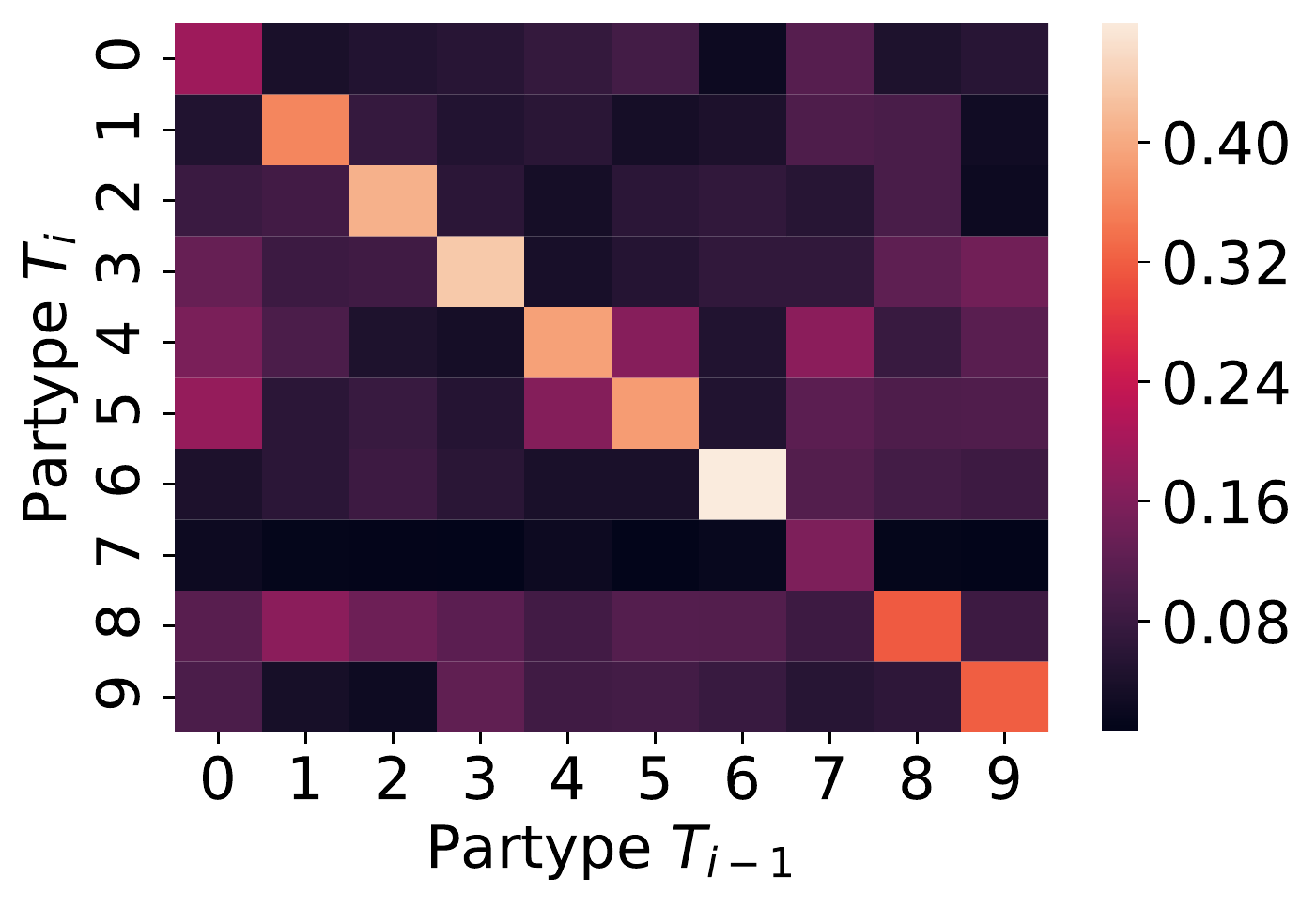}
    }
    \caption{Paragraph-Type Transition Matrices.}
    \label{fig:transition-matrices}
\end{figure}

We show in Figure \ref{fig:transition-matrices} the likelihood of transitioning from one paragraph-type to the next. The \textbf{KMeans} approach places a high weight on staying in the same paragraph-type, likely because it is picking up on document-specific words. Our model, on the other hand, learns a more broad paragraph-type transition behavior.

Next, we compare the distribution over paragraph-types learned by each model for the first paragraph. We expect apriori that the first paragraph in each article is most likely to be the lede paragraph, which shares characteristics. Thus, we'd expect a better model to learn a distribution over paragraph-types for the first paragraph that have less entropy. We show the results in Table \ref{tbl:entropy}. In this case, our model is outperformed by the baseline.

\begin{table}
\begin{tabular}{|l|r||l|r|}
\toprule
\multicolumn{2}{|c||}{\textbf{Top words,} $s=0$} & \multicolumn{2}{c|}{\textbf{Top words,} $s=1$}\\
\hline
\textbf{Word} &      $p(s=1)$ & \textbf{Word} &      $p(s=1)$  \\
\midrule
             cites &  0.27 &  diseases &  0.95 \\
         evidently &  0.27 &     scores &  0.95 \\
        accusation &  0.27 &     suburb &  0.94 \\
            verbal &  0.27 &   colombia &  0.94 \\
            assist &  0.27 &     pushed &  0.94 \\
          abstract &  0.29 &   depicted &  0.93 \\
\bottomrule
\end{tabular}
\caption{Vocabulary words with the highest likelihood of being associated with paragraph type ($s=0$) or document topic ($s=1$). Words with count greater than $50$ shown. }
\label{tbl:top-words-s-1}
\end{table}

\begin{table*}
\begin{tabular}{|p{.1\linewidth}|p{.35\linewidth}||p{.1\linewidth}|p{.35\linewidth}|}
\toprule
Par-Type & Top Topics & Par-Type & Top-Topics \\
\hline
0 &                    
\makecell{
law, today, million;\\
president, trial, crimes;\\
law, man, crimes\\
}
&
2 &        
\makecell{
department, general, trial;\\
united, percent, black;\\
million, trial, justice\\
}
 \\
 \hline
3 &        
\makecell{
president, states, today;\\
cases, evidence, charges;\\
office, million, prison\\
}
 &
4 &          
\makecell{
charges, department, percent;\\
today, public, law;\\
prison, criminal, today\\
}
 \\
 \hline
6 &      
\makecell{
charges, black, department;\\
prison, criminal, today;\\
president, states, today\\
}
&
7 &            
\makecell{
united, percent, black;\\
law, justice, according;\\
house, charges, crimes\\
}
 \\
 \hline
8 &     
\makecell{
million, trial, justice;\\
department, general, trial;\\
evidence, judge, criminal\\
}
 &
9 &  
\makecell{
prison, criminal, today;\\
president, district, charges;\\
department, general, trial\\
}
 \\
\bottomrule
\end{tabular}
\caption{Top topics associated with \textbf{selected} paragraph types. Top three topics are weighted by PMI.}
\label{tbl:partype-topics}
\end{table*}

\begin{table}[t]
    \centering
    \begin{tabular}{|c||c|c|}
    \hline
     Method & KMeans & Paragraph TM  \\
    \hline
    Entropy & 2.22 & 2.27\\
    \hline
    \end{tabular}
    \caption{Entropy over Paragraph Types for the First Paragraph of the article.}
    \label{tbl:entropy}
\end{table}

\subsection{Shortcomings and Extensions}

We note that our model has several shortcomings that may have limited our observations.

We do not model positional information about paragraphs. For instance, the ``lede'' paragraph might be more likely to be the first paragraph, and be more likely to be followed by a ``Background'' than a ``Secondary Event'' paragraph. Our model ignores this information. A possible extension could be the incorporation of $T_{i-1}$ into the conditional for $T_i$.

Additionally, we do not place a learnable prior on the switching variable, $s$ and simply fix $\gamma$. This might be too strong of a belief, and in future models we might want to place a $Beta$ distribution over $\gamma$. Further, we did not experiment with different settings for $\gamma$ because of the slow speed of our topic model, we simply ran the model with $\gamma = .7$.

Finally, and most importantly, we are only learning a single distribution over paragraph types, $T$. If, as noted by other researchers \cite{po2003news,khataei2013recommender,kabob,lang1989effects,dai2018fine}, there are several different article structures, it is logical to assume that $P_T$ is not single distribution over paragraph types, but rather multiple, leading us to introduce another variable for article type $A$, into $T$'s conditional.

\section{Generating Lede Paragraphs}

In this section, we describe our approach to generating lede paragraphs for crime stories. On a high level, we take crime reports generated by the Los Angeles Police Department (LAPD) and generate a set of templates for different crime reports. Our journalistic goal is to generate an output article for every hate crime committed in Los Angeles county.

\subsection{Data}
Our templates take as input police reports generated by the LAPD. The information provided in an example police report is shown in Figure \ref{subtbl:example-crime-report}. The LAPD publishes between 10-20 crime reports a week with the code \textbf{0903}, which indicate a \textbf{Hate Crime}. Each police report consists of a crime-type, a location, a list of codes, and other assorting metadata.

Each code maps to a description, enumerated in the LAPD's Consolidated Crime Analysis Database\footnote{\url{https://github.com/alex2awesome/lapd-hate-crimes/blob/master/data/CCADMANUAL\%202019\%20040319\%20-\%20excl.\%20SAR\%20codes.pdf}}. We show a sample of codes relevant to the example in Table \ref{subtbl:sample-crime-codes}.

We partner with the Crosstown organization, which is a journalistic organization based at University of Southern California's Annenberg School for Communication and Journalism. Student journalists at Annenberg have manually generated $\approx 300$ lede-sentences for hate crimes, starting January 1st, 2019. We use these ledes as training data for our system. A sample lede for the crime shown in Table \ref{tbl:sample-crime-code} is shown in Table \ref{subtbl:sample-lede}.

\begin{table}[t!]
    \centering
    \subfloat[Details provided by the LAPD crime report: crime type, location, and code. \label{subtbl:example-crime-report}]{
    \begin{tabular}{|p{.9\linewidth}|}
        \hline
        \underline{\textbf{Crime Report:}}\\
        \hline
        \underline{\textit{Crime Type}}: Simple Assault\\
        \hline
        \underline{\textit{Location}}: 3100 block of Wilshire Blvd., Los Angeles\\
        \hline
        \underline{\textit{LAPD Code Provided}}: \textbf{0359 1238 2057 1506 0305 0429 0903}\\
        \hline
    \end{tabular}
    }
    \\
    \subfloat[Annotations of the crime codes provided by the LAPD Consolidated Crime Analysis Database.
    \label{subtbl:sample-crime-codes}
    ]{
    \begin{tabular}{|p{.15\linewidth}|p{.7\linewidth}|}
    \hline
    \textbf{Code} & \textbf{LAPD Code Description}\\
    \hline
        \textbf{0359} & Suspect Behavior: Suspect made unusual statement \\
        \textbf{1238} & Victim details: Victim was handicapped \\
        \textbf{2057} & Victim details: Victim targeted based on disability \\
        \textbf{1506} & Bias: Physical disability \\
        \textbf{0305} & Attacks from rear \\
        \textbf{0429} & Victim knocked to ground \\
        \textbf{0903} & Hatred/Prejudice \\
    \hline
    \end{tabular}
    }
    \\
    \subfloat[A sample lede written by a student journalist.
    \label{subtbl:sample-lede}
    ]{
        \begin{tabular}{|p{.9\linewidth}|}
            \hline
            \underline{\textit{Desired lede}}: The suspect assaulted the 59-year-old black disabled male victim near the Wilshire/Vermont metro station, while shouting an "unusual" statement. \\
            \hline
        \end{tabular}
    }
    \\
    \subfloat[An example template for crime type \textbf{SIMPLE ASSAULT}.
    \label{subtbl:sample-template}
    ]{
    \begin{tabular}{|p{.9\linewidth}|}
    \hline
    The suspect, \textbf{<suspect description> <crime type verb> <victim description>} near \textbf{<location indicator>, <location description> <other details>}.\\
    \hline
    \end{tabular}
    }
    
    \subfloat[A sample of our rule-based template system. Errors shown in \textcolor{red}{\textbf{red}}.\label{subtbl:model-generated-output}]{
    \begin{tabular}{|p{.9\linewidth}|}
        \hline
        \underline{\textit{Generated lede}}: The suspect assaulted the 59-year-old black disabled male victim near the Wilshire/Vermont metro station, while shouting \textcolor{red}{\textbf{a}} unusual statement\textcolor{red}{\textbf{s}}.\\
        \hline
    \end{tabular}
    }
    \caption{An example of the data input that our template system takes, as well as the manual output by a student journalist, our template for crimes of this type, and the ledes we generate.}
    \label{tbl:sample-crime-code}
\end{table}

\subsection{Template Approach}
The size of the data is too small to learn an automatic generation of text. So, we focus on a semi-automatic way of learning and filling templates in this section. This attempt focuses on enabling more sophisticated learning in the future by enriching the parallel data for template and generated lede paragraphs. 

An analysis of the LAPD crime reports revealed an existence of natural clusters in the crime codes. All crime codes could be clustered into larger bins like suspect behavior, suspect details, victim details, bias etc. based on the description of these codes. 

The analysis also showed a variation in the design of lede paragraph based on the crime type. For example, a crime of type \textbf{ASSAULT-Hatred/Prejudice} will have focus on race/ethnicity of the victim and the suspect. This contrasts with, for example, crimes of type \textbf{VANDALISM - Misdemeanor}, which focus instead on the monetary damage as well.

Templates are hand-coded for each crime type based on a token and n-gram similarity across hand-written ledes. Hand-written lede paragraphs for a particular crime type are manually evaluated to get all spans of text occurring across all examples of the crime type. These spans are then joined together with ``<slot>''s  to create a template. The need for a <slot> is identified by examining mismatching text spans across ground truth lede paragraphs. We manually group crime-code clusters based on keyword match and identify sets of crime-codes applicable for each slot. A manual verification of these auto-generated templates is done to check for grammatical inconsistencies. An example of one such template for crime type = ``ASSAULT'' is shown in Table \ref{subtbl:sample-template}.

\textbf{Template Filling:}
A `slot-filling' approach has been used to auto-generate the lede paragraph from its corresponding template. A lookup is created for ``crime type verb'' slot of the template. Each crime type cluster has its details spread across multiple fields of LAPD crime data. These are joined together to fill the slot. For example, a ``victim description'' slot combines \textit{age, race/ethnicity, gender} with the word ``victim'' to generate ``the 59-year-old black disabled male victim''. For a few corner cases like slots with missing data fields in the LAPD crime data, the longest common span of text is found across ground truth lede paragraphs for that slot/crime cluster type. Other LAPD data fields like \textit{suspect behavior, weapon type, suspect history etc.} can also be incorporated in the generated lede paragraph during its manual verification. Manual evaluation is a required step for now, to check for grammatical consistencies and overall meaning of the generated lede.

\begin{table}
    \centering
    \subfloat[An example template for crime type \textbf{VANDALISM-DISMEANOR}.
    \label{subtbl:sample-template-v}
    ]{
    \begin{tabular}{|p{.9\linewidth}|}
    \hline
    The suspect, \textbf{<crime type verb>} property at \textbf{<location description>} resulting in damage of \textbf{<damage-value bracket>}.\\
    \hline
    \end{tabular}
    }
    \\
    \subfloat[The lede written by a student journalist.]{
        \begin{tabular}{|p{.9\linewidth}|}
             \hline
             \underline{\textit{Desired Lede}}: The suspect vandalized property at Marmion Apartments, resulting in damage of \$400 or less.
             \\ 
             \hline 
        \end{tabular}
    }\\
    \subfloat[The lede generated by our model. Errors shown in \textcolor{red}{\textbf{red}}.]{
        \begin{tabular}{|p{.9\linewidth}|}
             \hline
             \underline{\textit{Generated Lede}}: The suspect vandalized \textcolor{red}{\textbf{at}} Marmion \textcolor{red}{\textbf{Apt. causing a}} damage of \$400 or less.''
             \\
             \hline 
        \end{tabular}
    }
    \caption{Template for crime of this type, sample ledes generated both by our model and the student journalist for crime type \textbf{VANDALISM-DISMEANOR}.}
    \label{tbl:sample-lede-2}
\end{table}

\subsection{Experimental Results}
We evaluate the templates we generate based on n-gram overlap of generated lede paragraph with ``Crosstown'' ground truth paragraph, shown below: 
\begin{align*}
    Acc = \frac{|\text{n-gram}_{model} \cap \text{n-gram}_{test}|}{|\text{n-gram}_{test}|}
\end{align*}

This was evaluated across 31 crime types using bi-gram counts to yield an average overlap of 83.4\%. 

A manual evaluation of hand-written lede paragraphs with the generated ledes showed that the majority of differences are due to grammatical errors: the addition or subtraction of a few non-crime supporting part-of-speech tokens or special characters like dash, quotes etc in any token.

We show the output of our templates compared with the student templates that generated them. Comparison of generated lede with ground truth lede for crime type \textbf{``ASSAULT''} is shown in Table \ref{subtbl:sample-lede}, \ref{subtbl:model-generated-output}. As can be seen in the example, the errors are both grammatical, dealing with a plurality error in the lookup of code \textbf{0359}. Comparison of generated lede with ground truth lede for crime type \textbf{``VANDALISM-DISMEANOR''} is shown in Table \ref{tbl:sample-lede-2}. As can be seen, the errors are primarily phrasal: for example, the model does not insert ``property'', and it inserts ``causing a'' instead of ``resulting in''.

\subsection{Next Steps}

More work is needed to refine this pipeline. One possible direction is the incorporation of grammatical corrections. For our purposes, relatively minor grammar errors may likely be correctable via rule-based methods \cite{sidorovetal2013rule} or hidden markov models \cite{vidal1995grammatical}. Furthermore, recent research has shown human-level accuracy for grammar correction \cite{ge2018reaching} using sequence-to-sequence neural networks.

Additionally, we can address some grammatically-correct but phrasally suboptimal lede generations using style transfer. The goal of text-style is to produce semantically equivalent text segments adapted to linguistic patterns prefered by one domain. Style-transfer has been more robustly explored in visual learning, where early work by \newcite{gatys2015neural} exploited CNNs to split content and style, and progressing to more recent architectures \cite{jing2019neural}. Previous work has compared style transfer between scientific news and scientific articles, like this work compared the headlines of news articles with the titles of scientific articles \cite{fu2018style}. 

\section{Related Works}

\textit{\textbf{NOTE (2022-06-14): }The current work was done for a class project in Fall 2019 and is shown without changes. However, since then, several works of interest have since been published, which we will expand upon at the end of this section, for the curious reader. }

The study of structure in news articles has a long history in numerous fields. In the field of communications, \newcite{van1985structures} provides perhaps the most detailed ontology of tags for paragraph-level discourse in news. 

We summarize the author's main points: \textit{Summary} elements express overall messages given by the news article: the \textit{headline} and the \textit{lede} are subcomponents which introduce and summarize the main topic. \textit{Situation} elements are events that drive the main subject of the article, which occur in \textit{episodes} and their \textit{consequences}. The \textit{background} tells the context: \textit{previous events}, \textit{history elements}, or ideological frameworks that help the reader understand the subject. \textit{Conclusions} are analysis by the journalist. \textit{Verbal reactions} are comments solicited from external sources \cite{van1985structures}.

Communications scholars have identified various news structures that use these discourse items. The classical \textit{Inverted Pyramid} structure aims to tell events in terms of decreasing significance, starting with a \textit{Lede} and continuing through to the \textit{Situation}, \textit{Background} and \textit{Conclusion} \cite{po2003news}. The \textit{Martini Glass} is similar, but places a greater emphasis on a chronological ordering of events after the lede \cite{khataei2013recommender}, even if elements of the background comes first. The \textit{Kabob} structure also attempts to hook the reader with a lede, but focuses more on the story and main events before transitioning to consequences and conclusions \cite{kabob}. Finally, the \textit{Chronological Story} tells a purely chronological ordering of events and may not focus on broader points or conclusions \cite{lang1989effects}.

Researchers in computer science have formulated approaches to learning this structure. \newcite{yarlott2018identifying} has tagged a corpus of $50$ news articles from the ACE Phase 2 corpus according to the \newcite{van1985structures} scheme. Researchers test a set of baseline methods -- Logistic Regression, Random Forest Classifiers, Decision Trees over bag-of-word article representations -- to automatically classify paragraphs into one of the \newcite{van1985structures}'s paragraph-types. \newcite{dai2018fine} has tagged a separate corpus of $853$ news articles into one of the $4$ discourse structures mentioned above. These researchers test baseline methods -- support-vector machine over hand-crafted features -- as well to classify articles into one of these four structures. Neither of these authors achieve convincing performance in either their tasks, and we hypothesize that some primary hurdles that prevent them from achieving good performance lie in the simplicity of their models and their small datasets. 

Another approach to examining discourse structure in news articles is the use of unsupervised structure learned through heirarchical BiLSTM neural networks \cite{karimi2019learning}. However, authors in this case train their network do not provide any interpretation into the discourse patterns they learn. It is unclear whether their findings are applicable to our goals.

\textbf{Update: 2022-06-14.} First, and most importantly, is work by \newcite{choubey2020discourse} which released the first labeled dataset of news discourse structures. The authors used a modified Van Dijk schema to label 800 news documents. \newcite{spangher-etal-2021-multitask} built upon this work to improve classification accuracy and show how these discourse tags complemented other forms of discourse analysis.



\section{Conclusion}

We have shown an unsupervised approach to discourse tagging in news articles and generated templates for lede paragraphs. We hope our work can lead to machine-in-the-loop systems that help journalists work more efficiently.

\clearpage
\bibliography{acl2020}

\begin{thebibliography}{28}
\expandafter\ifx\csname natexlab\endcsname\relax\def\natexlab#1{#1}\fi

\bibitem[{Antunes and Hurley(1977)}]{antunes1977representation}
George~E Antunes and Patricia~A Hurley. 1977.
\newblock The representation of criminal events in houston's two daily
  newspapers.
\newblock \emph{Journalism Quarterly}, 54(4):756--760.

\bibitem[{Bamman et~al.(2013)Bamman, O’Connor, and
  Smith}]{bamman2013learning}
David Bamman, Brendan O’Connor, and Noah~A Smith. 2013.
\newblock Learning latent personas of film characters.
\newblock In \emph{Proceedings of the 51st Annual Meeting of the Association
  for Computational Linguistics (Volume 1: Long Papers)}, pages 352--361.

\bibitem[{Barthel(2018)}]{pewnewsadvertising}
Michael Barthel. 2018.
\newblock \href
  {https://www.pewresearch.org/fact-tank/2019/07/23/key-takeaways-state-of-the-news-media-2018/}
  {5 key takeaways about the state of the news media in 2018}.
\newblock \emph{Pew Research}.

\bibitem[{Blei et~al.(2003)Blei, Ng, and Jordan}]{blei2003latent}
David~M Blei, Andrew~Y Ng, and Michael~I Jordan. 2003.
\newblock Latent dirichlet allocation.
\newblock \emph{Journal of machine Learning research}, 3(Jan):993--1022.

\bibitem[{Chermak(1994)}]{chermak1994body}
Steven~M Chermak. 1994.
\newblock Body count news: How crime is presented in the news media.
\newblock \emph{Justice Quarterly}, 11(4):561--582.

\bibitem[{Choubey et~al.(2020)Choubey, Lee, Huang, and
  Wang}]{choubey2020discourse}
Prafulla~Kumar Choubey, Aaron Lee, Ruihong Huang, and Lu~Wang. 2020.
\newblock Discourse as a function of event: Profiling discourse structure in
  news articles around the main event.
\newblock In \emph{Proceedings of the 58th Annual Meeting of the Association
  for Computational Linguistics}.

\bibitem[{Cohen(1975)}]{cohen1975comparison}
Shari Cohen. 1975.
\newblock A comparison of crime coverage in detroit and atlanta newspapers.
\newblock \emph{Journalism Quarterly}, 52(4):726--730.

\bibitem[{Dai et~al.(2018)Dai, Taneja, and Huang}]{dai2018fine}
Zeyu Dai, Himanshu Taneja, and Ruihong Huang. 2018.
\newblock Fine-grained structure-based news genre categorization.
\newblock In \emph{Proceedings of the Workshop Events and Stories in the News
  2018}, pages 61--67.

\bibitem[{Freedman(2009)}]{freedman2009new}
Des Freedman. 2009.
\newblock ‘new’news environment.
\newblock \emph{New media, old news: Journalism and democracy in the digital
  age}, page~35.

\bibitem[{Fu et~al.(2018)Fu, Tan, Peng, Zhao, and Yan}]{fu2018style}
Zhenxin Fu, Xiaoye Tan, Nanyun Peng, Dongyan Zhao, and Rui Yan. 2018.
\newblock Style transfer in text: Exploration and evaluation.
\newblock In \emph{Thirty-Second AAAI Conference on Artificial Intelligence}.

\bibitem[{Gatys et~al.(2015)Gatys, Ecker, and Bethge}]{gatys2015neural}
Leon~A Gatys, Alexander~S Ecker, and Matthias Bethge. 2015.
\newblock A neural algorithm of artistic style.
\newblock \emph{arXiv preprint arXiv:1508.06576}.

\bibitem[{Ge et~al.(2018)Ge, Wei, and Zhou}]{ge2018reaching}
Tao Ge, Furu Wei, and Ming Zhou. 2018.
\newblock Reaching human-level performance in automatic grammatical error
  correction: An empirical study.
\newblock \emph{arXiv preprint arXiv:1807.01270}.

\bibitem[{Gilliam~Jr et~al.(1996)Gilliam~Jr, Iyengar, Simon, and
  Wright}]{gilliam1996crime}
Franklin~D Gilliam~Jr, Shanto Iyengar, Adam Simon, and Oliver Wright. 1996.
\newblock Crime in black and white: The violent, scary world of local news.
\newblock \emph{Harvard International Journal of press/politics}, 1(3):6--23.

\bibitem[{Jing et~al.(2019)Jing, Yang, Feng, Ye, Yu, and Song}]{jing2019neural}
Yongcheng Jing, Yezhou Yang, Zunlei Feng, Jingwen Ye, Yizhou Yu, and Mingli
  Song. 2019.
\newblock Neural style transfer: A review.
\newblock \emph{IEEE transactions on visualization and computer graphics}.

\bibitem[{Karimi and Tang(2019)}]{karimi2019learning}
Hamid Karimi and Jiliang Tang. 2019.
\newblock Learning hierarchical discourse-level structure for fake news
  detection.
\newblock \emph{arXiv preprint arXiv:1903.07389}.

\bibitem[{Katz(1987)}]{katz1987makes}
Jack Katz. 1987.
\newblock What makes crimenews'?
\newblock \emph{Media, Culture \& Society}, 9(1):47--75.

\bibitem[{Khataei and Lau(2013)}]{khataei2013recommender}
Amirsam Khataei and Diana Lau. 2013.
\newblock Recommender narrative visualization.
\newblock In \emph{Proceedings of the 2013 Conference of the Center for
  Advanced Studies on Collaborative Research}, pages 415--421. IBM Corp.

\bibitem[{Lang(1989)}]{lang1989effects}
Annie Lang. 1989.
\newblock Effects of chronological presentation of information on processing
  and memory for broadcast news.
\newblock \emph{Journal of Broadcasting \& Electronic Media}, 33(4):441--452.

\bibitem[{Mann(2018)}]{crimestat}
Stephen Mann. 2018.
\newblock \href {https://blog.oup.com/2018/04/crime-news-media-america/} {Crime
  and the media in america}.
\newblock \emph{Harvard International Journal of press/politics}.

\bibitem[{Marsh(1991)}]{marsh1991comparative}
Harry~L Marsh. 1991.
\newblock A comparative analysis of crime coverage in newspapers in the united
  states and other countries from 1960--1989: A review of the literature.
\newblock \emph{Journal of Criminal Justice}, 19(1):67--79.

\bibitem[{Oliveres(2017)}]{kabob}
Victoria Oliveres. 2017.
\newblock \href
  {https://medium.com/narrative-from-linear-media-to-interactive-media/structuring-a-local-immersive-feature-aimed-at-the-world-f4a24abb0b8e}
  {Structuring a local immersive feature aimed at the world}.
\newblock \emph{Medium}.

\bibitem[{Po{\"{}}~ttker(2003)}]{po2003news}
Horst Po{\"{}}~ttker. 2003.
\newblock News and its communicative quality: The inverted pyramid—when and
  why did it appear?
\newblock \emph{Journalism Studies}, 4(4):501--511.

\bibitem[{Schildkraut(2017)}]{schildkraut2017crime}
Jaclyn Schildkraut. 2017.
\newblock Crime news in newspapers.
\newblock In \emph{Oxford Research Encyclopedia of Criminology and Criminal
  Justice}.

\bibitem[{Sidorov et~al.(2013)Sidorov, Gupta, Tozer, Catala, Catena, and
  Fuentes}]{sidorovetal2013rule}
Grigori Sidorov, Anubhav Gupta, Martin Tozer, Dolors Catala, Angels Catena, and
  Sandrine Fuentes. 2013.
\newblock \href {https://www.aclweb.org/anthology/W13-3613} {Rule-based system
  for automatic grammar correction using syntactic n-grams for {E}nglish
  language learning ({L}2)}.
\newblock In \emph{Proceedings of the Seventeenth Conference on Computational
  Natural Language Learning: Shared Task}, pages 96--101, Sofia, Bulgaria.
  Association for Computational Linguistics.

\bibitem[{Spangher et~al.(2021)Spangher, May, Shiang, and
  Deng}]{spangher-etal-2021-multitask}
Alexander Spangher, Jonathan May, Sz-Rung Shiang, and Lingjia Deng. 2021.
\newblock \href {https://doi.org/10.18653/v1/2021.emnlp-main.40} {Multitask
  semi-supervised learning for class-imbalanced discourse classification}.
\newblock In \emph{Proceedings of the 2021 Conference on Empirical Methods in
  Natural Language Processing}, pages 498--517, Online and Punta Cana,
  Dominican Republic. Association for Computational Linguistics.

\bibitem[{Van~Dijk(1985)}]{van1985structures}
Teun~A Van~Dijk. 1985.
\newblock Structures of news in the press.
\newblock \emph{Discourse and communication: New approaches to the analysis of
  mass media discourse and communication}, 10:69.

\bibitem[{Vidal et~al.(1995)Vidal, Casacuberta, and
  Garc{\'\i}a}]{vidal1995grammatical}
Enrique Vidal, Francisco Casacuberta, and Pedro Garc{\'\i}a. 1995.
\newblock Grammatical inference and automatic speech recognition.
\newblock In \emph{Speech Recognition and Coding}, pages 174--191. Springer.

\bibitem[{Yarlott et~al.(2018)Yarlott, Cornelio, Gao, and
  Finlayson}]{yarlott2018identifying}
W~Victor Yarlott, Cristina Cornelio, Tian Gao, and Mark Finlayson. 2018.
\newblock Identifying the discourse function of news article paragraphs.
\newblock In \emph{Proceedings of the Workshop Events and Stories in the News
  2018}, pages 25--33.

\end{thebibliography}
\bibliographystyle{acl_natbib}

\end{document}